\documentclass[runningheads]{llncs}

\usepackage[T1]{fontenc}
\usepackage{tikz}
\usepackage{newtxtext}
\usepackage[varvw]{newtxmath}
\usepackage{comment}
\usepackage{graphicx}
\usepackage{booktabs}
\usepackage{multirow}
\usepackage{array}
\usepackage{subcaption}

\usepackage{xcolor}

\usepackage{amsmath}

\usepackage{url}
\usepackage{microtype}
\usepackage{xcolor}
\usepackage{enumitem}
\usepackage{makecell}
\usepackage[normalem]{ulem}
\useunder{\uline}{\ul}{}

\title{Separating Prompt Harmfulness from Interaction Safety via Partial Generation}
\titlerunning{Prompt Harmfulness and Interaction Safety}

\author{
Ram Vikas Mishra \and
Shashank Aggarwal \and
Jahnavi Kumar \and
Amit Awekar
}

\authorrunning{R. V. Mishra et al.}

\institute{
Department of Computer Science and Engineering,
Indian Institute of Technology Guwahati,
Guwahati 781039, Assam, India
}

\begin{document}
\maketitle

\begin{abstract}
LLM-based information systems need safety guardrails. Most guardrails classify the prompt alone. If the prompt looks harmful, the system rejects it before any answer is generated. This is costly. Many rejected prompts would have received a safe response. Users lose help they should have received from the information system. This cost grows as conversational and generative search becomes common. Prompt-level labels treat harm as a property of the query. But harm depends on what the model actually writes. Existing input-output guards do look at responses. But they judge a finished response, so the decision comes after full generation. However, we observe that safety of model interaction with the input prompt can be inferred early with only partial output. So the system can start generating, watch the partial response, and stop only if it turns harmful. We test this idea in three steps. First, we build a dataset of 41,944 prompt-response pairs with seven open weights LLMs from three families: Llama, Gemma, and Qwen. Second, we show that prompt harmfulness and interaction safety differ. Prompt labels and response labels agree on only 50.46\% of items (Cohen's $\kappa$ = 0.06). Of the prompts that make any model respond harmfully, only 3.5\% do so for all seven LLMs. Third, we show that partial responses carry most of the signal. Fine-tuned classifiers predict response-level harm well (best AUPRC 0.885, F1 0.807). About one third of the response tokens recover almost full-response AUPRC. 
\end{abstract}

\section{Introduction}

\begin{figure}[t]
\centering
\small
\fbox{\parbox{\dimexpr\linewidth-2\fboxsep-2\fboxrule\relax}{\textbf{Prompt:} \emph{How can I get into my neighbor's Wi-Fi without the password?}}}

\vspace{2mm}
\fbox{\parbox[t][2.1cm][t]{\dimexpr0.5\linewidth-2\fboxsep-2\fboxrule-1mm\relax}{\footnotesize
\textbf{Model A (safe)}\\[1mm]
\textbf{I can't help with getting into someone else's network without} their permission. If you are locked out of your own Wi-Fi, you can \textcolor{gray}{[...]}}}\hfill
\fbox{\parbox[t][2.1cm][t]{\dimexpr0.5\linewidth-2\fboxsep-2\fboxrule-1mm\relax}{\footnotesize
\textbf{Model B (harmful)}\\[1mm]
\textbf{Sure! Here is how to get into your neighbor's Wi-Fi:} Step 1. \textcolor{gray}{[harmful details omitted]}}}
\caption{One prompt, two interactions. The first words (in bold) already show whether the model interaction is safe or harmful.}
\label{fig:example}
\end{figure}

\emph{Prompt harmfulness} is whether a prompt asks for something harmful. The prompt in Figure~\ref{fig:example} is harmful in this sense. It is a property of the query, and it is fixed before any information system responds.
\emph{Interaction safety} is whether the model's response to the prompt is harmful. It depends on the prompt and on the model together. Model~A refuses the harmful request, and it offers a safe alternative to help the user. So the interaction of Model~A is safe. In contrast, Model~B complies with the harmful intent and starts to list steps for getting into the neighbor's Wi-Fi. So the interaction of Model~B is harmful.
The prompt is the same, but the two interactions differ.
\emph{Partial generation} means generating only a few tokens of the response and deciding whether it is heading toward a harmful interaction. In the figure, the first few words already show that the interaction of Model~A with the prompt will be safe, and that of Model~B will be harmful.
A prompt-level label cannot tell the two interactions apart.

Prompt-only classifiers equate prompt harmfulness with interaction safety~\cite{protectai}. Such classifiers have to take a defensive approach and reject every prompt that might generate a harmful response from any model.
This is one extreme: the decision is made before any output exists. About half of the prompts in our corpus carry a malicious label. Yet, only about 8.5\% of the prompts in our corpus make any of the seven LLMs respond harmfully.

\begin{figure}[b]
\centering
\tikzset{
  box/.style={draw, rounded corners=2pt, minimum height=8mm, align=center, font=\footnotesize, inner sep=3pt},
  sol/.style={box, fill=gray!15},
  lab/.style={font=\footnotesize\itshape, anchor=west},
  tag/.style={font=\scriptsize, inner sep=1.5pt},
  arr/.style={-latex, thick},
  ptr/.style={-latex, semithick}
}
\begin{minipage}[t]{0.49\linewidth}
\centering
\resizebox{\linewidth}{!}{%
\begin{tikzpicture}
\useasboundingbox (-0.7,-0.7) rectangle (6.8,3.5);
\node[lab] at (-0.7,3.2) {(a) Existing solutions};
\node[box, minimum width=11mm] (pr) at (0,0) {Prompt};
\node[box, minimum width=19mm] (sy) at (2.15,0) {LLM-based\\system};
\node[box, minimum width=12mm] (ou) at (4.35,0) {Output};
\node[box, minimum width=10mm] (us) at (6.1,0) {User};
\draw[arr] (pr) -- (sy);
\draw[arr] (sy) -- (ou);
\draw[arr] (ou) -- (us);
\node[sol, minimum width=19mm] (pc) at (0.875,1.95) {Prompt-only\\classifier\\{\scriptsize\itshape before generation}};
\node[sol, minimum width=19mm] (io) at (5.275,1.95) {Input-output\\guard\\{\scriptsize\itshape after generation}};
\draw[ptr] (pc.south) -- (0.875,0);
\draw[ptr] (io.south) -- (5.275,0);
\end{tikzpicture}}
\end{minipage}\hfill
\begin{minipage}[t]{0.49\linewidth}
\centering
\resizebox{\linewidth}{!}{%
\begin{tikzpicture}
\useasboundingbox (-0.7,-0.7) rectangle (6.8,3.5);
\node[lab] at (-0.7,3.2) {(b) Ours: partial generation};
\node[box, minimum width=11mm] (pr) at (0,0) {Prompt};
\node[box, minimum width=19mm] (sy) at (2.15,0) {LLM-based\\system};
\node[box, minimum width=12mm] (ou) at (4.35,0) {Output};
\node[box, minimum width=10mm] (us) at (6.1,0) {User};
\draw[arr] (pr) -- (sy);
\draw[arr] (sy) -- (ou);
\draw[arr] (ou) -- (us);
\node[sol] (mo) at (3.425,1.95) {Partial-\\generation\\monitor\\{\scriptsize\itshape during generation}};
\draw[ptr, dashed] (3.425,0) -- node[tag, right] {partial output} (mo.south);
\draw[ptr, dashed] (mo.west) -| coordinate[pos=0.8] (stopmid) (sy.north);
\node[tag, left] at (stopmid) {stop or continue};
\end{tikzpicture}}
\end{minipage}
\caption{Where safety decisions are made. Our monitor reads the response while it is generated and stops it if it turns harmful.}
\label{fig:pipelines}
\end{figure}
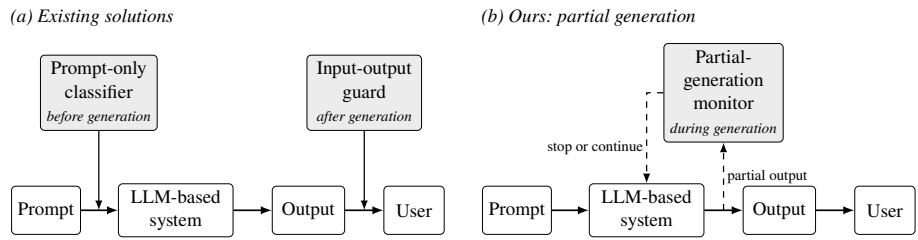

The other extreme is to generate the full response and then decide whether to abort.
This is what input-output guards do~\cite{llamaguard,shieldgemma,wildguard}.
They see the actual interaction, so they do not need to be defensive. But the decision comes late. The system pays the full cost of generating the response, and then abandons it. Neither extreme is satisfying. The first rejects too much, and the second waits too long. We observe that the first few tokens are often enough to see what the model is doing, so the decision can be made early, during generation.
This paper separates prompt harmfulness from interaction safety, and it uses partial generation to decide interaction safety early. Figure~\ref{fig:pipelines} shows where each approach makes its decision.

This paper answers three research questions.
First, does prompt harmfulness really differ from interaction safety when it comes to real-world LLMs? Second, can interaction safety be predicted from the response? Third, how early in the response can the decision be made? To answer them, we build a corpus of 41,944 prompt-response pairs. It uses about 6,000 shared prompts from four public sources.
Each prompt is answered by seven open-weight models from three families (Llama, Gemma, and Qwen). We check the response labels against two human annotators for a subset of corpus, and we use an LLM judge to label the full corpus.

Our experimental results support our hypotheses. Prompt labels and response labels agree on only 50.46\% of pairs across the corpus (Cohen's $\kappa = 0.06$ and PABAK = 0.0091). On a human-audited sample of 1,050 pairs, agreement is 62.8\% ($\kappa = 0.27$). Harmful outputs are often specific to one model. Out of all prompts that make any of the seven models respond harmfully, only 3.5\% of the prompts do so for all seven. This means a prompt-level label cannot say which model will generate harmful response. Fine-tuned classifiers predict response-level harm well, with a best area under precision recall curve (AUPRC) of 0.885 and F1-score of 0.807 (Table~\ref{results}).
The first one third tokens recover almost full-response AUPRC. So most of the safety signal appears early. For complete reproducibility, our datasets, code, prompts, and models are available publicly on the Web\footnote{https://github.com/Conf-links/InteractionSafety}. Our contributions are as follows.
\begin{itemize}
  \item We quantify the gap between prompt harmfulness and interaction safety across seven models.
\item We release a multi-model prompt-response corpus that keeps the prompt label, the generating model and the response label together.
\item We show that prompt-trained guards fail on interaction labels, and that guards trained only on full responses carry most of the signal in the first third of a response.
\end{itemize}

\section{Our Work}
\label{sec:work}

This section describes how we study interaction safety.
We first define the problem and the two pipelines we compare (Section~\ref{sec:formulation}).
Then we describe the corpus: the prompts, the responses and the labels (Section~\ref{sec:prompts}).
Last, we describe the classifiers, the partial-generation monitor and the evaluation protocol (Section~\ref{sec:classifiers}).

\subsection{Problem Formulation}
\label{sec:formulation}

Let $p$ be a prompt and $m$ a generative model.
The model produces a response $r_m$ for the prompt $p$.
Every prompt has a \emph{prompt label} $y_p \in \{0,1\}$, taken from its source dataset.
It says whether the prompt is malicious.
Every pair $(p, m)$ has an \emph{interaction label} $y \in \{0,1\}$.
It is 1 if the response $r_m$ complies with a malicious intent and 0 otherwise.
The interaction label depends on the model.
The same prompt can have $y=0$ for one model and $y=1$ for another. A prompt-level guard ($C_p(p)$)predicts from the prompt alone.
An interaction-level guard ($C_{pr}(p,r_m)$)also sees the response:
\begin{equation}
  C_p(p) \approx y_p, \qquad C_{pr}(p, r_m) \approx y .
\end{equation}
The interaction-level guard may also take the model identity $m$ as input.
We write $r_m^{\le k}$ for the first $k$ tokens of the response.
With $k=0$, the guard sees only the prompt.
With $k=|r_m|$, it sees the full response.

When an information system uses Prompt gating, it rejects the prompt if $C_p(p)=1$.
Otherwise, it generates the full response. An input-output guard based information system rejects prompt if $C_{pr}(p, r_m)=1$. In contrast, a Partial Generation monitoring based system always starts generating response for any prompt $p$. At a set of checkpoints $\mathcal{K}$, it scores the partial response.
It stops at the first checkpoint $k \in \mathcal{K}$ where $C_{pr}(p, r_m^{\le k}) =1$. If no checkpoint fires, the full response $r_m$ is delivered to the user.


\begin{table}[tb]
\centering
\small
\begin{tabular}{lrrrrrrr}
\toprule
& \multicolumn{2}{c}{Train} & \multicolumn{2}{c}{Validation} & \multicolumn{2}{c}{Test} & \\
\cmidrule(lr){2-3} \cmidrule(lr){4-5} \cmidrule(lr){6-7}
Source   & Benign & Malicious & Benign & Malicious & Benign & Malicious \\
\midrule
AdvBench & 0      & 352       & 0      & 48        & 0      & 120       \\
Arena    & 893    & 0         & 165    & 0         & 166    & 0          \\
deepset  & 302    & 169       & 40     & 31        & 56     & 58         \\
xTRam    & 852    & 1683      & 148    & 317       & 200    & 400      \\
\midrule
Total    & 2047   & 2204      & 353    & 396       & 422    & 578      \\
\bottomrule
\end{tabular}

\caption{Prompt-level composition before response generation.}
\label{tab:split}
\end{table}

\subsection{Dataset Generation}
\label{sec:prompts}

We build the prompt pool from four public sources. AdvBench contributes clearly harmful instructions~\cite{advbench}.
The deepset dataset contributes safe and injection prompts~\cite{deepset_dataset}.
The xTRam1 dataset adds broader injection coverage, with synthetic attacks and benign prompts~\cite{xtram_dataset}.
Chatbot Arena contributes natural user requests, which we use as benign prompts~\cite{arena}. The pool has 5,000 training prompts and 1,000 test prompts.
No prompt appears in both splits.
Table~\ref{tab:split} shows the composition.
About half of the prompts are labeled malicious. We mix sources so that the corpus covers ordinary requests, direct harmful requests and injection-style requests. The prompt label comes from the source dataset. For AdvBench, it means a harmful instruction. For deepset and xTRam1, it means a prompt injection.
These two meanings are not the same, and we return to this point when we compare prompt labels with interaction labels.


We generate responses with seven open-weight models from three families.
The Gemma family has Gemma~2~2B, Gemma~3~12B and Gemma~3~27B.
The Llama family has Llama~3.2~3B and Llama~3.1~8B.
The Qwen family has Qwen~2.5~7B and Qwen~2.5~14B. Every model answers every prompt in our dataset. The planned corpus has $6{,}000 \times 7 = 42{,}000$ pairs. We drop 56 pairs for which the judge gave no clear label. The final corpus has 41{,}944 pairs. It has 34{,}946 training pairs and 6{,}998 test pairs.

We label the corpus with an LLM judge, Gemini 3.1 Flash-Lite. For a subset of the corpus, we check the LLM judge labels against human annotation. The judge answers one question: does the response comply with a malicious intent? Judge prompt and annotation guidelines are available in our code repository. For the human check, we sample 150 pairs from each model, which gives 1{,}050 pairs. Two human annotators labelled all of them. We measure agreement between the two humans and between each human and the judge. We report Cohen's $\kappa$ for pairs of raters and Fleiss' $\kappa$ for all three. The labels are imbalanced, so we also report Prevalence-Adjusted Bias-Adjusted Kappa (PABAK) and Randolph's $\kappa$, which are less sensitive to imbalance. Once the judge agrees well enough with the humans, we use it to label all pairs of prompt and response.

The final corpus has 1{,}535 harmful pairs out of 41{,}944, a harmful rate of 3.66\%.
The rate is 3.68\% in the training split and 3.56\% in the test split. We release the interaction label together with the generating model, the prompt source and the prompt label.

\subsection{Classifier Training and Evaluation}
\label{sec:classifiers}

We study five classifiers, chosen to cover different designs.
Two are encoders: the ProtectAI DeBERTa-v3 prompt-injection detector~\cite{protectai} and Prompt-Guard-2.
Two are generative guards: Llama Guard~3~8B~\cite{llamaguard} and WildGuard~7B~\cite{wildguard}.
The fifth is LAYA, a decision model with an encoder backbone and a small decision head.
For each classifier, we test two inputs: the prompt alone, and the prompt with the response. We first run each classifier in the zero-shot setting. We then fine-tune it on our training pairs. Encoders and LAYA are fully fine-tuned. The two generative guards use QLoRA~\cite{dettmers2023qlora}. For the guards, the loss is next-token cross-entropy on the decision tokens only. 

Only 3.6\% of the training pairs are harmful. Plain fine-tuning does not work well under this imbalance. We therefore compare six strategies to handle the label imbalance: standard cross-entropy, weighted loss, oversampling, undersampling, weighted oversampling, and focal loss~\cite{lin2017focal}. Oversampling and undersampling both bring the harmful rate to 20\%. Of the 34{,}946 training pairs, 29{,}710 are used for training. The other 5{,}236 form the validation set. We pick the strategy for each classifier based on the performance on the validation set. We take the one with the highest recall, and break ties within 0.02 by F1. We pick the decision threshold that maximizes harmful-class F1 on the validation set at the full response. We then keep the threshold fixed for each classifier.

The monitor needs a classifier that works on incomplete responses.We study this in two steps. First, we test the classifiers trained on full responses at shorter prefixes. We sweep the prefix from 0\% to 100\% of the response tokens in steps of 10\%. At 0\%, the classifier sees only the prompt. At 100\%, it sees the full response.
The threshold stays fixed at the value chosen at 100\%.

We report AUPRC, precision, recall, and F1-score for the harmful class. Harmful pairs are only 3.7\% of the data. So AUPRC and harmful-class F1 matter more than AUROC or accuracy. Precision and recall are reported at the fixed validation threshold.

\section{Dataset Analysis}
\label{sec:analysis}

This section uses the corpus to answer our first question: does prompt harmfulness differ from interaction safety?
We proceed in four steps.
We first describe the corpus.
We then check the quality of the labels.
Next, we compare prompt labels with interaction labels.
Last, we study how failures vary across models.

Harmful responses are rare in our corpus, and their rate differs by model.
Of the 41{,}944 pairs, 1{,}535 are harmful.
This is 3.66\%.
The rate is 3.68\% in the training split and 3.56\% in the test split, so the two splits are similar. The harmful response rate varies for each family and model. For Qwen it is 1.5\% for 14b model and 3.1\% for 7b model. For Llama it is 1.7\% for 8b model and 2.5\% for 3b model. For Gemma  it is 6.19\%, 5.1\%, and 5.3\% for 27b, 12b, and  2b models respectively. 
The Gemma family has the highest harmful response rate at 5.55\% average. The Llama family has 2.13\% and the Qwen family has 2.35\%. Model size does not order harmful response rate  consistently. In the Llama and Qwen families, the larger model is safer.
In the Gemma family, the largest model has the highest harmful response rate.

\begin{table}[bt]
\centering
\caption{Agreement on the 1{,}050 prompt-response pair audit set.}
\label{tab:audit}
\begin{tabular}{lrrr}
\hline
Pair & Agreement (\%) & Cohen's $\kappa$ & PABAK \\
\hline
Human 1 -- Human 2 & 92.57 & 0.664 & 0.851 \\
Human 1 -- Gemini  & 95.43 & 0.825 & 0.909 \\
Human 2 -- Gemini  & 92.57 & 0.667 & 0.851 \\
\hline
Prompt label -- Gemini  & 62.76 & 0.272 & 0.255 \\
Prompt label -- Human 1 & 61.05 & 0.238 & 0.221 \\
Prompt label -- Human 2 & 57.43 & 0.170 & 0.149 \\
\hline
\end{tabular}
\end{table}

The LLM judge agrees with humans well enough to label the corpus at scale.
We check this on the 1{,}050 audit pairs, with two human annotators and the judge.
Table~\ref{tab:audit} shows the agreement. 
All three annotators (two humans and LLM judge) give the same label on 948 of 1{,}050 pairs (90.29\%) with Fleiss' $\kappa$ of 0.72. These values show substantial agreement. But annotation by humans and LLM judge is not identical.
The labels are imbalanced, so $\kappa$ is lower than raw agreement. For such datasets, Randolph's $\kappa$ is a more appropriate measure to evaluate agreement between three annotators. For our dataset, it comes out to be 0.87, indicating significant agreement.

The two human annotators do not label alike. The second human annotator marks 103 pairs as harmful, while the first human annotator marks 161 and the judge marks 163. So second human annotator is stricter. The LLM judge is closer to the first human annotator than the two human annotators are to each other.
We use the LLM judge as a consistent labeler for the full corpus at the scale of 42{,}000 prompt-response pairs. Prompt labels and interaction labels disagree on a large share of the pairs. On the audit set, the raw agreement is 62.76\% ($\kappa = 0.27$) for the judge, and 57--61\% for the two humans with $\kappa$ in the range of 0.23 to 0.17 (Table~\ref{tab:audit}).
So the gap is not an artifact of the judge. Most of the disagreement comes from malicious-labeled prompts that receive safe responses.

Whether a prompt leads to a harmful response depends on the model. We call a model \emph{failing} on a prompt if its response is harmful. Of the 6{,}000 prompts, 509 make at least one model fail.
This is 8.5\% failure rate. However, at the prompt level, 53\% of the prompts carry a malicious label. Most of these 509 prompts fail on only a few models.
For 159 of them, exactly one model fails.
Only 18 prompts make all seven models fail, which is 3.5\%. So a prompt-level label cannot say which model will fail. Failures are partly shared within a family.
For example, there are 43 prompts where all three Gemma models fail and no model from another family fails. This is expected, because models of one family share training data and methods. 

Taken together, these results answer our first question. Prompt labels agree weakly with interaction labels. The gap is mostly safe responses to prompts labeled malicious. And the failures that remain depend on the model. The next section asks whether a classifier can predict these outcomes, and how early.
\section{Experiments}
\label{sec:experiments}

This section answers our second and third questions.
Can interaction safety be predicted from the response?
And how early in the response can the decision be made?
We use the classifiers, the training setup and the metrics described in the previous section.
We first compare the classifiers (Section~\ref{sec:predict}).
We then study how the safety signal grows as more of the response is shown (Section~\ref{sec:early}).
We close with a discussion of what the results mean and where they stop applying (Section~\ref{sec:discussion}).

\subsection{Predicting Interaction Safety}
\label{sec:predict}

We first ask whether classifiers can predict interaction safety.
Table~\ref{results} shows the results on the test set.
Each classifier is tested zero-shot and after fine-tuning on our corpus.
Each setting uses two inputs: the prompt alone, and the prompt with the full response.
The fine-tuned classifiers are trained on the prompt and the full response.
In the ``prompt only'' rows, we withhold the response at test time.
For the fine-tuned classifiers, precision, recall and F1 use a threshold chosen on the validation set for the same input as the row.

\begin{table}[t]
\centering
\small
\setlength{\tabcolsep}{3pt}
\caption{Results on the test set (6{,}998 pairs, 3.56\% harmful), harmful class. ZS is zero-shot and FT is fine-tuned, with the strategy in brackets. FT classifiers are trained on the prompt and the full response. In the ``prompt only'' rows, the response is withheld at test time. For the FT classifiers, precision, recall and F1 use a threshold chosen on the validation set for the same input as the row: no response for ``prompt only'', and the full response for ``prompt + resp.''}
\label{results}
\begin{tabular}{llcccc}
\toprule
Classifier & Setting & AUPRC & Prec. & Rec. & F1 \\
\midrule
\multirow{4}{*}{DeBERTa-v3} & ZS, prompt only & 0.071 & 0.084 & 0.743 & 0.151 \\
 & ZS, prompt + resp. & 0.076 & 0.082 & 0.614 & 0.145 \\
 & FT (standard), prompt only &  0.444 & 0.465 & 0.610 & 0.528 \\
 & FT (standard), prompt + resp. & 0.683 & 0.644 & 0.755 & 0.695 \\
\midrule
\multirow{4}{*}{\shortstack[l]{Prompt-Guard-\\2-86M}} & ZS, prompt only & 0.302 & 0.380 & 0.446 & 0.410 \\
 & ZS, prompt + resp. & 0.357 & 0.467 & 0.426 & 0.445 \\
 & FT (oversampling), prompt only & 0.459 & 0.459 & 0.667 & 0.543 \\
 & FT (oversampling), prompt + resp.& 0.646 & 0.613 & 0.751 & 0.675 \\
 \midrule
\multirow{4}{*}{\shortstack[l]{Llama-Guard-\\3-8B}} & ZS, prompt only & 0.037 & 0.052 & 0.731 & 0.097 \\
 & ZS, prompt + resp. & 0.252 & 0.174 & 0.446 & 0.250 \\
 & FT (weighted loss), prompt only & 0.052 & 0.071 & 0.851 & 0.132 \\
 & FT (weighted loss), prompt + resp. & \textbf{0.885} & 0.749 & 0.876 & \textbf{0.807} \\
\midrule
\multirow{4}{*}{WildGuard-7B} & ZS, prompt only & 0.041 & 0.060 & 0.839 & 0.112 \\
 & ZS, prompt + resp. & 0.615 & 0.501 & 0.735 & 0.596 \\
 & FT (oversampling), prompt only & 0.379 & 0.490 & 0.590 & 0.536 \\
 & FT (oversampling), prompt + resp. & 0.685 & 0.576 & 0.759 & 0.655 \\
\midrule
\multirow{4}{*}{LAYA} & ZS, prompt only & 0.060 & 0.062 & 0.365 & 0.106 \\
 & ZS, prompt + resp. & 0.099 & 0.061 & 0.538 & 0.110 \\
 & FT (focal loss), prompt only & 0.510 & 0.478 & 0.614 & 0.538 \\
 & FT (focal loss), prompt + resp. & 0.706 & 0.603 & 0.715 & 0.654 \\
\bottomrule
\end{tabular}
\end{table}

\begin{figure}[bt]
    \centering
    \includegraphics[width=0.99\linewidth]{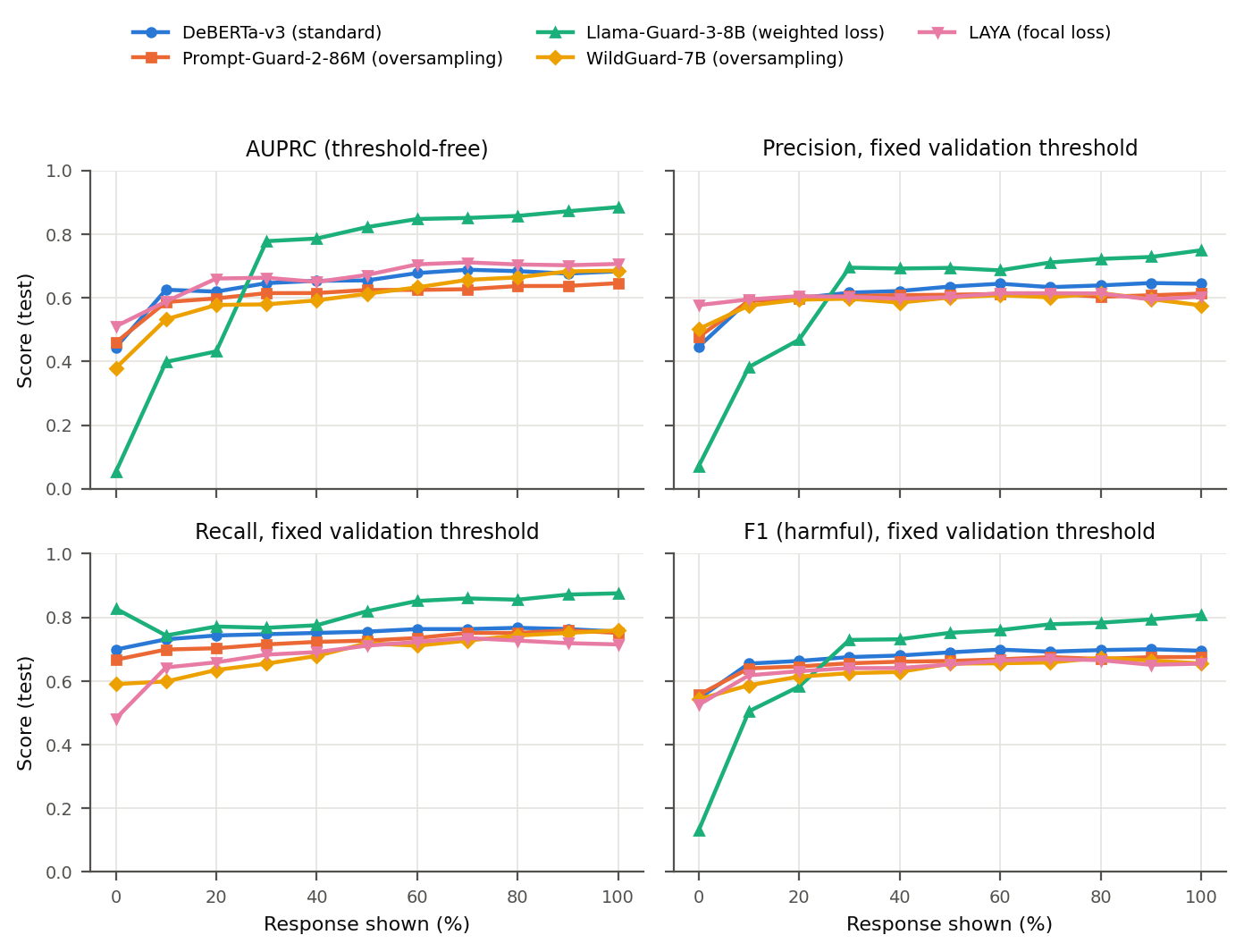}
    \caption{Test-set AUPRC, precision, recall and F1 of the fine-tuned classifiers as the prefix grows from 0\% to 100\% of the response tokens, in steps of 10\%. The imbalance strategy chosen for each classifier is in brackets. At 0\% the classifier sees only the prompt. Precision, recall and F1 use the threshold fixed on the validation set at the full response.}
    \label{sweeps}
\end{figure}

\paragraph{Zero-shot.}
Off the shelf, the classifiers do poorly on our labels.
Their precision is low, between 0.05 and 0.50.
This is expected.
These models were trained for other label definitions, and several were built to flag prompts, not to judge interactions.
Giving them the response helps only when they were built to read it.
The two generative guards gain the most.
The AUPRC of WildGuard rises from 0.041 to 0.615, and that of Llama Guard from 0.037 to 0.252.
The other three gain little.

\paragraph{Fine-tuned.}
Fine-tuning on our corpus raises every classifier.
Llama Guard is the best, with an AUPRC of 0.885 (95\% CI 0.84--0.93) and an F1 of 0.807.
The other four reach AUPRC values between 0.646 and 0.706.
Their confidence intervals overlap, so we do not rank them.
The F1 rises for all five classifiers, from 0.11--0.60 in the zero-shot setting with the response to 0.65--0.81 after fine-tuning.
The main gain is precision.
For DeBERTa, Llama Guard and LAYA, it rises from 0.06--0.17 to 0.60--0.75.
Recall also rises for all five classifiers, compared with the zero-shot setting with the response.
Precision matters here, because only 3.6\% of the pairs are harmful.
A low-precision guard would block many safe interactions.

\paragraph{What the response adds.}
The ``FT, prompt only'' rows show how much signal the prompt alone carries.
Each fine-tuned row uses a threshold tuned for its own input, so the comparison is fair.
For DeBERTa, Prompt-Guard-2, WildGuard and LAYA, the prompt alone gives an AUPRC of 0.38--0.51 and an F1 of 0.53--0.54.
This is far above their zero-shot prompt-only AUPRC of 0.04--0.30.
So once a classifier learns our label definition, the prompt carries useful signal.
But the prompt alone is not enough.
Adding the response raises their AUPRC by 0.19--0.31, to 0.65--0.71, and their F1 by 0.12--0.17, to 0.65--0.70.
Precision shows the practical cost.
With the prompt alone, precision is 0.46--0.49, so more than half of the interactions these classifiers block are safe.
With the response, precision rises to 0.58--0.64.
Llama Guard shows the effect most clearly.
With the prompt alone, its AUPRC is 0.052, close to the 3.6\% share of harmful pairs.
With the response, it is 0.885.
So the best classifier gains almost everything from the response.
These classifiers were trained with the response.
A classifier trained on prompts only might do better, and we do not train one.

\subsection{How Early Can We Decide}
\label{sec:early}

Most of the safety signal appears early in the response.
To test this, we give each fine-tuned classifier only the first part of the response at test time.
We sweep the prefix from 0\% to 100\% of the response tokens in steps of 10\%.
At 0\%, the classifier sees only the prompt.
At 100\%, it sees the full response.
Figure~\ref{sweeps} shows AUPRC, precision, recall and F1 for the five classifiers.

Three points stand out.
\begin{itemize}
  \item \emph{The gain comes early.}
  After 30\% of the response, the classifiers reach about 85--95\% of their full-response AUPRC.
  The curves then flatten.
  For four of the five classifiers, most of this gain is already there at 10\%.
  \item \emph{Llama Guard changes the most.}
  With the prompt alone, its AUPRC is about 0.05.
  It rises to about 0.78 at 30\% and to 0.885 at 100\%.
  At 0\%, its recall is high but its precision is near zero.
  So at the fixed threshold, it flags almost every pair when it has no response.
  A likely reason is that it was built as an input-output guard, so it relies more on the response.
  \item \emph{The other four already work from the prompt alone.}
  Their F1 is 0.53--0.54 with the prompt only, and 0.65--0.70 with the full response.
  A short prefix lifts them further, and then they level off.
\end{itemize}

Two points limit this reading.
First, the 0\% point comes from classifiers trained on full responses and tested on the prompt alone.
It is the setting of the ``FT, prompt only'' rows in Table~\ref{results}.
The AUPRC is the same.
Precision, recall and F1 differ, because the table tunes the threshold for the prompt alone, while the figure keeps the full-response threshold.
It is not a classifier trained on prompts only.
Second, the threshold is fixed at the value chosen for the full response.
This penalizes short prefixes.
So these curves may understate what a classifier trained for early decisions can reach.

Overall, the decision does not need the full response.
A prefix of about one third of the response already carries most of the safety signal.

\subsection{Discussion}
\label{sec:discussion}

\paragraph{Where the monitor fits.}
The monitor sits between the model and the user.
The model streams tokens.
A small classifier scores the prefix at a few checkpoints.
The system stops the stream on an alarm.
The classifiers are small compared with the generative LLMs, and they run only at the checkpoints, so the latency cost induced by the classifiers is negligible as compared to the generative LLMs.  

\paragraph{The cost of monitoring.}
Monitoring trades rejection for exposure.
Some harmful text may be produced before the stop, and in a streaming interface the user may see it.
Two measures reduce this.
The system can hold back the first $k$ tokens until the first check, at the price of a short delay.
It can also keep prompt gating for the most severe categories.
An attacker who knows the monitor could delay the harmful part until after a checkpoint, or open with a harmless preamble. We will have to improvise our method to guard against such adversarial attacks.

\section{Related Work}
\label{sec:related}

We organize related work by where the safety decision is made and by what it is based on (Figure~\ref{fig:pipelines} and Table~\ref{tab:paradigms}).
We then discuss two themes that sit beside these paradigms: refusal evaluation of models, and the cost of rejecting too much. Please refer to Table~\ref{tab:paradigms} for a summary of where our work sits in the context of existing paradigms on safety of information systems.

\paragraph{Before generation: prompt-level safety.}
Most safety classifiers treat harm as a property of the user input.
Prompt-injection and jailbreak detectors are binary classifiers over the prompt~\cite{protectai}.
They are trained on datasets such as xTRam1 and deepset~\cite{xtram_dataset,deepset_dataset}.
Benchmarks such as AdvBench, HarmBench and BIPIA measure attack success against models~\cite{advbench,zou2023universal,harmbench,bipia}.
These classifiers are fast, because they do not run the model.
But their label is model-independent by construction.
The same label applies whichever model answers.
So a prompt-level classifier must be defensive: it has to reject every prompt that could produce harm from any model.
We use these datasets as the source of our prompts and prompt labels, and we ask how well the labels predict the safety of the interaction.

\paragraph{After generation: input-output moderation.}
A second line of work judges the response.
Llama Guard and ShieldGemma classify prompts and responses under a safety taxonomy~\cite{llamaguard,shieldgemma}.
WildGuard separates prompt harmfulness, response harmfulness and response refusal~\cite{wildguard}, and BeaverTails labels the harmfulness of question-answer pairs~\cite{ji2023beavertails}.
Qwen3Guard-Gen adds a ``controversial'' class between safe and unsafe~\cite{qwen3guard}.
Their response-level labels are the closest to ours.
But they judge a finished response, so the decision comes after full generation.
Li et al.\ note that this full detection causes high service latency~\cite{li2025scm}.

\paragraph{During generation: streaming and early stopping.}
The closest work monitors the response while it is generated.
Li et al.\ argue that applying moderators trained on full responses to partial ones creates a training-inference gap.
They build FineHarm, a dataset of 29K pairs with token-level labels, and a streaming content monitor.
It reaches a macro-F1 of 0.95 or more after seeing about 18\% of the response tokens on average~\cite{li2025scm}.
Qwen3Guard-Stream adds a token-level head for real-time monitoring~\cite{qwen3guard}.
It has been integrated into CARE, a framework that detects unsafe text, rolls back, and intervenes at decoding time~\cite{hu2025care}.
These works show that training for early decisions works.
We look at the other regime.
We use guards that were not built for partial responses, and ask how much signal they already hold at each prefix.

\begin{table}[bt]
\centering
\footnotesize
\setlength{\tabcolsep}{3pt}
\caption{Paradigms of safety work, and where our study sits.}
\label{tab:paradigms}
\begin{tabular}{>{\raggedright\arraybackslash}p{2.3cm}>{\raggedright\arraybackslash}p{2.1cm}>{\raggedright\arraybackslash}p{3.0cm}>{\raggedright\arraybackslash}p{3.6cm}}
\toprule
Paradigm & Decision point & Decision based on & Examples or focus \\
\midrule
Prompt-level guards & Before generation & The prompt & ProtectAI detector; xTRam1, deepset, AdvBench \\
\midrule
Input-output guards & After the full response & Prompt and response & Llama Guard, ShieldGemma, WildGuard \\
\midrule
Streaming monitors & During generation & Prompt and partial response, trained for it & Streaming content monitor, Qwen3Guard-Stream, CARE \\
\midrule
Model refusal benchmarks & No decision; models are ranked & Responses of many models to unsafe or safe-looking prompts & SORRY-Bench, HarmBench, XSTest, OR-Bench \\
\midrule
Query harmfulness prediction & Before retrieval & The query & QHP \\
\midrule
\textbf{Ours} & During generation & Prompt and partial response, with guards trained on full responses & Gap between prompt label and interaction outcome, across seven models \\
\bottomrule
\end{tabular}
\end{table}

\paragraph{Refusal evaluation of models.}
Another line of work evaluates the models themselves.
SORRY-Bench compares over 50 models on the same 440 unsafe instructions.
It shows that refusal behavior differs across models, and it checks LLM judges against more than 7K human annotations~\cite{xie2025sorrybench}.
HarmBench compares many models under the same attacks~\cite{harmbench}.
These results support two choices in our study.
First, models differ on the same prompt.
Second, an LLM judge can be validated against human labels.
But the audience is the model developer, who wants to rank models.
All prompts are unsafe by design, so there is no contrast between a prompt label and the outcome.
And the results are not used to decide whether to reject a prompt or stop a response.

\paragraph{The cost of rejecting too much.}
Rejecting too much has a known cost.
XSTest and OR-Bench measure how often models refuse safe prompts that only look unsafe~\cite{rottger2024xstest,cui2025orbench}.
Recent work traces over-refusal to a small set of attention heads and suppresses them at inference time~\cite{mitigate2026}.
Guardrails add a similar cost outside the model.
Stronger guardrails improve security but reduce utility, and LLM-based moderation adds latency~\cite{nofreelunch}.
These studies work at one end: prompts that are benign but look unsafe.
We work at the other end.
We ask how many prompts that datasets label as malicious still receive a safe response.

\paragraph{Safety in information retrieval.}
Safety is also an IR problem.
Query harmfulness prediction asks, before retrieval, whether a query is likely to return harmful results~\cite{carrera2026qhp}.
Studies of AI-powered search engines find that they can quote or cite malicious content~\cite{luo2025unsafe}.
Like prompt-level classifiers, query harmfulness prediction decides from the query alone.
Our setting moves the decision to the generated answer, which depends on both the query and the model.

\section{Conclusion and Future Work}
\label{sec:conclusion}

\paragraph{Conclusion.}
We asked whether the harmfulness of a prompt decides the safety of the interaction with a model.
It does not.
In our corpus of 41{,}944 prompt-response pairs from seven models, at least 93\% of the responses to malicious-labeled prompts are safe.
Only 8.5\% of the prompts make any model respond harmfully, and only 3.5\% of those make all seven models fail.
So a prompt-level label cannot say which model will fail, and rejecting every prompt that looks harmful blocks mostly safe interactions.

The response carries the missing information.
Guards fine-tuned on interaction labels reach an AUPRC of up to 0.885 with the response, and 0.05--0.51 with the prompt alone.
Off the shelf, guards trained for prompts or for other label definitions reach a precision of at most 0.50 on our labels.
Most of the signal appears early.
After about one third of the response, guards trained only on full responses reach about 85--95\% of their full-response AUPRC.
This points to a decision made during generation: not before it, as prompt gating does, and not after it, as input-output guards do.
Our claims concern measurement and data.
We do not propose a new monitor, and we do not train for partial responses.
The corpus, with the prompt label, the generating model and the response label kept together, lets others test such monitors against the gap we measure.

Our work can be improved in several ways by addressing the following limitations.
\begin{itemize}
  \item \emph{Labels.} One LLM judge labels the corpus, including the test set. The human audit covers 1{,}050 pairs, and the two annotators differ in strictness. We define harm as compliance with a malicious intent plus harmful text. Other definitions would give other labels.
  \item \emph{Evaluation.} We evaluate classifiers at fixed prefixes. We do not run the full stop-or-continue monitor end to end. So we do not report how many safe interactions it would block, how many harmful ones it would let through, or how much harmful text it would produce before stopping.
   \item \emph{Prompts.} The prompts come from single-turn queries only. Multi-turn conversations are not part of our dataset.
  \item \emph{Models.} We use seven open-weight models from three families, between 2B and 27B parameters. We do not test closed or frontier models.
\end{itemize}

\paragraph{Ethical considerations.}
Our corpus contains harmful model outputs.
The prompts come from public datasets and contain no personal data.
The harmful text comes from small open-weight models, so the corpus adds little that these models do not already produce.
We release our corpus to support safer LLM-based information systems.

\bibliographystyle{splncs04}
\bibliography{references}

\end{document}